\documentclass[letterpaper]{article} % DO NOT CHANGE THIS
\usepackage{graphicx}
\usepackage{subcaption}
\usepackage{aaai25}  % DO NOT CHANGE THIS
\usepackage[table,xcdraw]{xcolor}
\usepackage{algorithm}
\usepackage{algpseudocode}
\usepackage{array}
\usepackage{amsmath}
\usepackage{amssymb}
\usepackage{times}  % DO NOT CHANGE THIS
\usepackage{helvet}  % DO NOT CHANGE THIS
\usepackage{courier}  % DO NOT CHANGE THIS
\usepackage[hyphens]{url}  % DO NOT CHANGE THIS
\usepackage{graphicx} % DO NOT CHANGE THIS
\usepackage{natbib}  % DO NOT CHANGE THIS AND DO NOT ADD ANY OPTIONS TO IT
\usepackage{caption} % DO NOT CHANGE THIS AND DO NOT ADD ANY OPTIONS TO IT
\usepackage{algorithm}
\usepackage{newfloat}
\usepackage{listings}
\DeclareCaptionStyle{ruled}{labelfont=normalfont,labelsep=colon,strut=off} % DO NOT CHANGE THIS
\floatstyle{ruled}
\newfloat{listing}{tb}{lst}{}
\floatname{listing}{Listing}

\title{ICU-TSB: A Benchmark for Temporal Patient Representation Learning for Unsupervised Stratification into Patient Cohorts}

\author{
    Dimitrios Proios\textsuperscript{1}, 
    Alban Bornet\textsuperscript{1}, 
    Anthony Yazdani\textsuperscript{1}, 
    Jose F Rodrigues Jr\textsuperscript{2},
    Douglas Teodoro\textsuperscript{1}
    \\ \{dimitrios.proios, 
    alban.bornet, 
    anthony.yazdani, 
    douglas.teodoro\}@unige.ch, junio@icmc.usp.br  \\
}

\affiliations{
    \textsuperscript{1}Department of Radiology and Medical Informatics, Faculty of Medicine, University of Geneva, Geneva, Switzerland; \\ \textsuperscript{2}Institute of Mathematics and Computer Science, University of Sao Paulo, Sao Carlos, Brazil.
}
\usepackage{bibentry}
\begin{document}

\maketitle

\begin{abstract}
Patient stratification—identifying clinically meaningful subgroups—is essential for advancing personalized medicine through improved diagnostics and treatment strategies. Electronic health records (EHRs), particularly those from intensive care units (ICUs), contain rich temporal clinical data that can be leveraged for this purpose. In this work, we introduce ICU-TSB (Temporal Stratification Benchmark), the first comprehensive benchmark for evaluating patient stratification based on temporal patient representation learning using three publicly available ICU EHR datasets. A key contribution of our benchmark is a novel hierarchical evaluation framework utilizing disease taxonomies to measure the alignment of discovered clusters with clinically validated disease groupings. In our experiments with ICU-TSB, we compared statistical methods and several recurrent neural networks, including LSTM and GRU, for their ability to generate effective patient representations for subsequent clustering of patient trajectories. Our results demonstrate that temporal representation learning can rediscover clinically meaningful patient cohorts; nevertheless, it remains a challenging task, with v-measuring varying from up to 0.46 at the top level of the taxonomy to up to 0.40 at the lowest level. To further enhance the practical utility of our findings, we also evaluate multiple strategies for assigning interpretable labels to the identified clusters. The experiments and benchmark are fully reproducible and available at  \url{https://github.com/ds4dh/CBMS2025stratification}.
\end{abstract}

% Uncomment the following to link t

\section{Introduction}
The identification of clinically relevant subtypes, known as patient stratification, plays a role in personalized medicine by improving clinical decision-making, refining diagnostic markers, and reducing healthcare costs \cite{ref4}. With the increasing availability of electronic health records (EHRs), machine learning methods have thrived in uncovering hidden patient subgroups and predicting health outcomes. For instance, \citet{carr2021longitudinal} demonstrated that recurrent neural networks (RNNs) could effectively cluster EHR data, capturing both patient trajectories and long-term health outcomes.

Despite significant advances in supervised learning for EHR-based predictions, temporal patient representation learning and unsupervised patient stratification remain underexplored. Supervised approaches rely on labeled data, limiting their generalizability to produce new cohorts and reach more healthcare institutions. In contrast, patient representation learning can extract meaningful patient representations from raw EHRs to support unsupervised stratification without requiring labeled outcomes. Recent works have proposed encoding patient trajectories into vector embeddings using sequence-based models such as Word2Vec \cite{jaume2022cluster, bornet2025comparing} and Graph Transformers \cite{choi_learning_2020}. However, these methods face challenges in scaling across large, heterogeneous datasets and handling the complexity of clinical ontologies such as ICD (International Classification of Diseases).

A major limitation in the field is the lack of standardized benchmarks for evaluating unsupervised patient stratification models. Existing research has focused mainly on supervised classification tasks, such as mortality prediction and disease phenotyping \cite{haru, van_de_water_yet_2024}. In contrast, benchmarking patient stratification in an unsupervised setting remains difficult due to the absence of ground-truth labels. Researchers have leveraged hierarchical rediscovery techniques to address this gap, which validate clustering methods by comparing their outputs to existing medical taxonomies \cite{ref4, ref5, ref2, ref3}.

This work proposes a reproducible evaluation framework for patient stratification using multicentric ICU datasets and hierarchical disease taxonomies. Specifically, we make the following contributions:

\begin{itemize}
    \item Benchmarking: We introduce ICU-TSB, a benchmark to evaluate patient representation learning in the task of unsupervised stratification on open-source ICU datasets.
    \item Patient representation learning: We compare statistical methods with deep learning architectures (e.g., LSTMs and GRUs) for generating patient embeddings.
    \item Hierarchical rediscovery: We formulate a patient clustering task based on ICD and CCS taxonomies, assessing model performance in rediscovering known disease categories.
    \item Cluster interpretability: We evaluate multiple label assignment strategies (e.g., centroid, medoid, and majority vote) to improve interpretability. % in unsupervised clustering.
\end{itemize}    

By leveraging open-source ICU data and patient representation learning, we establish a foundation for future research in scalable, explainable patient stratification.

\section{Methods}

% \begin{figure}
% \renewcommand{\arraystretch}{1.3}
% \begin{tabular}{|>{\centering\arraybackslash}m{2.4cm}|>{\centering\arraybackslash}m{2.4cm}|>{\centering\arraybackslash}m{2.4cm}|}
% \hline
% \rowcolor{gray!20} \textbf{1. Pre-\newline processing} & \textbf{2. Representa-\newline tion Learning} & \textbf{3. Unsupervised\newline Clustering} \\ \hline
% % \rowcolor{gray!20} \textbf{1. Pre-processing} & \textbf{2. Representation Learning} & \textbf{3. Unsupervised Clustering} \\ \hline
% \cellcolor{yellow!20} 
%     \raggedright
% \begin{itemize}
%     \item Categorical Encoding
%     \item Feature dimension Imputation
%     \item Split train/test/stay
%     \item Grouping by stay
% \end{itemize} & 
% \cellcolor{yellow!20}
% \begin{itemize}
%     \item Temporal dimension imputation
%     \item STAT
%     \item LSTM
%     \item GRU
% \end{itemize} & 
% \cellcolor{yellow!20}
% \begin{itemize}
%     \item Stratification  
%     \item Label naming assignment
%     \item Hierarchical clustering 
%     \item $t$-SNE
%     \item $k$-Means 
% \end{itemize} \\ \hline

% \end{tabular}

%     \caption{Overview of architecture pipeline. }
%     \label{fig:pipeline}
% \end{figure}

\subsection{Dataset description}

For ICU-TSB, we used data from three publicly available ICU repositories \cite{miiv, sic, eicu}, encompassing multiple ICU units and covering diseases from nearly all ICD chapters, albeit with varying frequencies. The feature space consists of 114 variables, including 108 hourly-sampled time series and 6 static features (i.e., non-temporal attributes), as summarized in Table \ref{table:dsum}. These features capture demographics, vital signs, mechanical support indicators, and clinical assessments, representing critical ICU monitoring data for assessing patient clinical condition.

\begin{figure}[!ht]
    \includegraphics[scale=0.34]{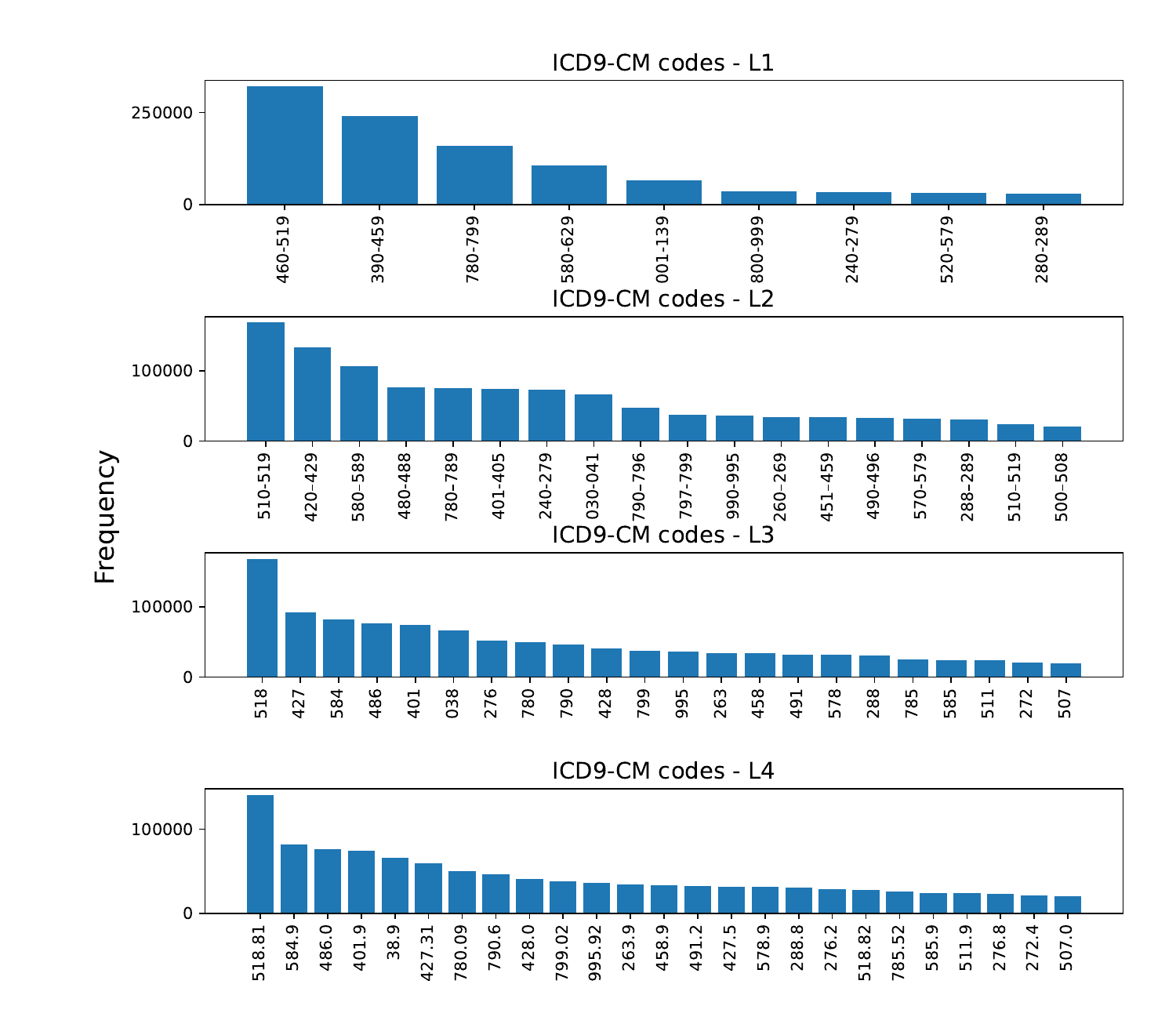}
% \begin{figure}\includegraphics[scale=0.22]{imgs/fig2.png}
    \caption{Frequency histogram of top 25 labels across datasets and levels of ICD-9-CM hierarchical taxonomy.}
    \label{fig:icdfreq}
\end{figure}

ICD-9-CM and ICD-10 codes are available for each ICU stay in three datasets, enabling disease classification. Given the irregular and skewed distribution of diagnoses, we selected the top 25 most frequent ICD-10 labels, mapping them to ICD-9-CM where necessary. This preparation ensures a sufficient sample size and statistical power while accounting for the long-tailed nature of disease prevalence across all levels of the the hierarchy, as illustrated in Figure \ref{fig:icdfreq}. To standardize data collection across datasets, we applied an hourly time granularity to all time-series features.

\begin{table}
    \centering
    \begin{tabular}{l|r|r|r|l}
        \textbf{Dataset} & 
        \textbf{\small Features} & 
        \textbf{Stays} & 
        \textbf{Codes} & 
        \textbf{Time unit}\\
        \hline
        eICU & 114 & 173,109 & 919 & Hour \\
        \hline
        % HiRID & 100 & 33,905 & - & Minute \\
        % \hline
        MIMIC-IV & 113 & 73,175 & 37,690 & Hour \\
        \hline
        SiC & 86 & 27,386 & 2,169 & Minute \\
    \end{tabular}
        
    \caption{Descriptive statistical summary across datasets.}
    \label{table:dsum}
\end{table}

\subsection{Data preprocessing}
We used the ricu R package \cite{ricu}, an open-source library that allows the application of temporal, value-based filtering and normalization. Furthermore, we performed preprocessing steps, including ordinal and one-hot encoding of categorical variables.
% We applied feature and time imputations to account for missing values and align feature values across datasets. 
We applied feature and time imputation accounting for missing values and align feature values across datasets. 
For example, eICU contains boolean flags of mortality for all patients, whereas other datasets  include only one of the two states. 
Last, we applied feature Robust Scaler \cite{sklearn} normalization using the training sets after grouping data by ICU stay. 

\subsection{Temporal patient representation learning}
We employed statistical and deep learning baselines to generate temporal embeddings from ICU data in the derived dataset. We used one statistical (STAT) and two recurrent neural networks (RNN) methods to generate patient representations over this dataset.

We trained the two deep learning baselines using autoregressive unidirectional LSTM and GRU models based on the work of \citet{SCEHR} and \citet{haru}. Three distinct LSTM-based models, one per dataset, were trained and optimized for the autoregressive task of inferring the feature values of the next time step. In contrast, the STAT baseline does not involve any learning process. For patient representations, we utilized the hidden state of the last timestep. In the case of STAT, we used concatenated statistical moments across time windows to create ICU stay representations, as described in the work of \citet{proios2023leveraging}.

We used a single channel accepting a 108-dimensional feature vector for deep learning baselines. In addition, we experimented with distinct LSTM cells for each feature. Subsequently, we replaced the LSTM cells with gated recurrent unit (GRU) cells and repeated the experiments.

\begin{figure}
    \centering
    \includegraphics[scale=0.27]{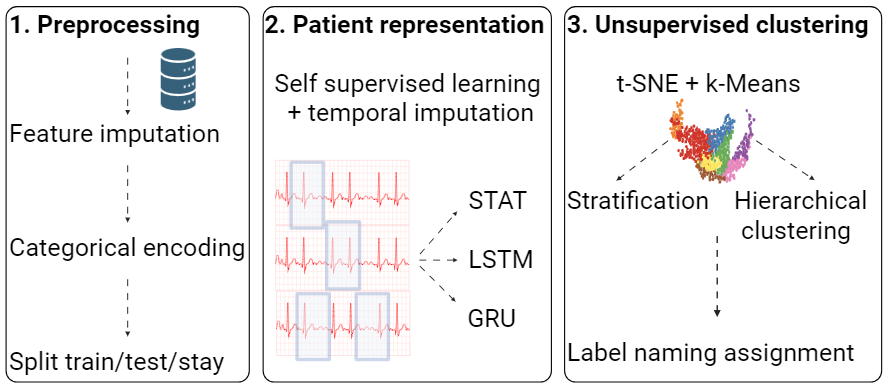}
    \caption{Overview of the architecture pipeline.}
    \label{fig:pipeline}
\end{figure}
We employed the generated embeddings as input for training unsupervised and semi-supervised clustering models to evaluate the capacity to identify patient cohorts. Our overall pipeline is summarized in Figure \ref{fig:pipeline}.

\subsection{ICD hierarchical tree levels}   
\label{sec:icd_foundation}  
The ICD is a hierarchical coding system that classifies diseases across the four levels $L_i$, $i \in \{1,2,3,4\}$. Let $\mathcal{L}_i$ be the set of codes at level $L_i$. Let $x_p \in \mathbb{R}^d $ represent the embedding vector for patient $p$, where $d$ is the dimensionality of the embedding space. These embeddings are generated by the self-supervised models (e.g., LSTM, GRU, STAT) to encapsulate the patient's clinical trajectory information. Each patient $p \in \mathcal{P}$ is associated with $y_{p,i} \in \mathcal{L}_i$, where $y_{p,i}$ represents the disease code at a specific level in the ICD hierarchy.  

\begin{figure}
    \centering
    \includegraphics[scale=0.45]{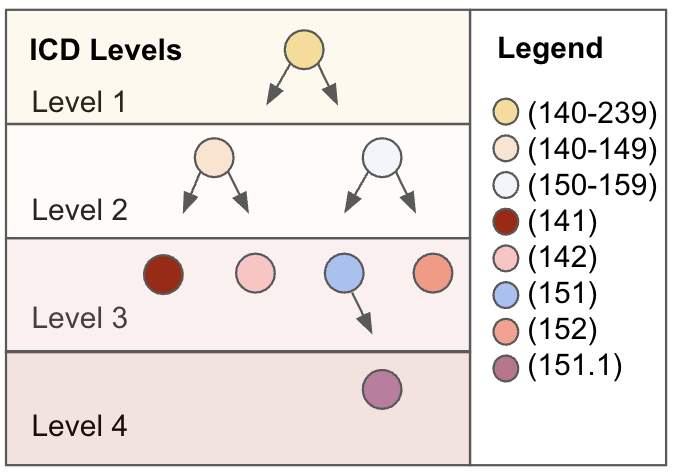}
    \caption{ICD-9-CM hierarchical structure for a subset of the Neoplasms chapter.}
    \label{fig:icdhieararchy}
\end{figure}

The mapping \(f_i: \mathcal{L}_i \to \mathcal{L}_{i-1}\) ensures that each code \(c_i \in \mathcal{L}_i\) has exactly one parent in $\mathcal{L}_{i-1}$. This hierarchical structure defines a rooted tree \(L = (V, E)\), visualized in Figure~\ref{fig:icdhieararchy} showing the four levels and code relationships, where
\begin{equation}
V = \mathcal{L}_1 \cup \mathcal{L}_2 \cup \mathcal{L}_3 \cup \mathcal{L}_4,
\end{equation}  
\begin{equation}
E = \{(c_i, c_{i+1}) \mid c_{i+1} \in \mathcal{L}_{i+1},\, c_i = f_{i+1}(c_{i+1})\}.
\end{equation}

\subsection{Patient stratification}

In this task, we formulate patient stratification as an unsupervised problem. We use the ICD hierarchy to extrinsically evaluate a model's ability to align with $\mathcal{L}_i$. Formally, a clustering model  (e.g., $k$-Means) maps each patient \(p\) to a cluster $K^{(i)}_j$, where \(K^{(i)}\) represents the set of clusters produced for $L_i$. 
In this setting, we evaluate the derived clusters with respect to the $j$-th ICD code at the $i$-th level. In the ideal case, each cluster \(K^{(i)}_j\) comprises a single code $j$, i.e., composed of all patients with identical $y_{p,i}$.

\subsection{Hierarchy rediscovery}
In this problem formulation, we define an iterative clustering problem, to approximate a clustering model from broad to specific levels. Let 
\begin{equation}
g^{(i+1)}: (\mathcal{P}, K^{(i)})  \to K^{(i+1)},
\end{equation}
where \(K^{(i)}\) represents the hierarchical cluster assignments of $\mathcal{P}$ at level $i$, using the prior set of clusters $K^{(i)}$ with $K^0=\emptyset$. The process has the following steps:

\begin{enumerate}
    \item \textit{Initial clustering at level \( L_1 \):}  
    Cluster all patients \( \mathcal{P} \) into \( k \) clusters corresponding to the broadest ICD categories in \( L_1 \):
    \begin{equation}
    K^{(1)} = g^{(1)}(\mathcal{P}, \emptyset).
    \end{equation}
    
    \item \textit{Iterative refinement for subsequent levels:}  
    For each subsequent level \( L_{i+1} \), we refine each cluster from the previous level by clustering the subset of patients belonging to that cluster:
    \begin{equation}
    K^{(i+1)} = g^{(i+1)}(\mathcal{P}, K^{(i)}).
    \end{equation}
\end{enumerate}
This hierarchical approach mirrors the ICD tree structure by partitioning data into increasingly granular subcategories.

\subsection{Cluster label assignment}
\label{sec:cluster_label_assignment}

In this task, we assign a label to each cluster \(K_j^{(i)} \in K^{(i)}\) using the true labels \(y_{p,i}\) of the patients belonging to that cluster. Although labels \(y_{p,i}\) exist for each patient \(p\), they remain unseen during the clustering process, preserving the unsupervised nature of the task. Instead, labels are utilized post-clustering to evaluate the quality of the cluster assignments in a transductive classification setting. Let 
\begin{equation}
    \mathcal{P}_{K_j^{(i)}} = \{ p \in  K_j^{(i)}\}
\end{equation}
denote the set of patients in cluster \(K_j^{(i)}\).
We define the cluster labeling function \(\ell\) as follows:

\begin{equation}
    \ell(K_j^{(i)}) = y_{p^*,i},
\end{equation}
where \(\ell(K_j^{(i)})\) assigns to cluster \(K_j^{(i)}\) the most representative label $y_{p,i}$, that is, $y_{p^*,i}$, among the available labels in $\mathcal{P}_{K_j^{(i)}}$. We consider three strategies to determine \(p^*\), using only the true labels of the training set:

\textit{1. Centroid-based.}  
Assign the label of the patient whose embedding is closest to the cluster centroid:
\begin{equation}
\mu_j = \frac{1}{|\mathcal{P}_{K_j^{(i)}}|} \sum_{p \in \mathcal{P}_{K_j^{(i)}}} x_p,
\end{equation}

where
\begin{equation}
p^* = \text{argmin}_{p \in \mathcal{P}_{K_j^{(i)}}} \| x_p - \mu_j \|_2.
\end{equation}
   
\textit{2. Medoid-based.}  
Assign the label of the patient whose embedding minimizes the total distance to all other embeddings in the cluster:
\begin{equation}
p^*=\text{argmin}_{p \in \mathcal{P}_{K_j^{(i)}}} \sum_{q \in \mathcal{P}_{K_j^{(i)}}} \| x_p - x_q \|_2.
\end{equation}

\textit{3. Majority-vote.}  
Assign the most frequent true label among patients in \(K_j^{(i)}\). Formally, 
\begin{equation}
% \ell(K_j^{(i)}) = 
p^*= \text{argmax}_{c \in \mathcal{L}_i} \sum_{p \in \mathcal{P}_{K_j^{(i)}}} \delta\left( y_{p,i} = c \right),
\end{equation}
where c represents a candidate label from the set of possible labels and $\delta$ the  indicator function summing the number of patients with a particular label defined as:
\begin{equation}
\delta\left( y_{p,i} = c \right) =
\begin{cases}
    1, & \text{if } y_{p,i} = c, \\
    0, & \text{otherwise}. 
\end{cases}
\end{equation}
These strategies assign a single representative label \(\ell(K_j^{(i)})\) to each cluster \(K_j^{(i)}\), enabling a consistent comparison of the cluster assignments against the true labels \(y_{p,i}\).

For evaluation, we employ v-measure \cite{rosenberg2007vmeasure}, adjusted mutual information (AMI) \cite{vinh2010information}, and accuracy. We optimize hyperparameters (including number of clusters and \(t\)-SNE parameters) using optuna \cite{optuna} with 50 trials for each formulation of the problem.

\subsection{Experimental setting}
In our experiments, we compare compare distinct representation learning and statistical methods—specifically STAT, LSTM-based, and GRU-based models in three problem formulations: clustering, label assignment, and hierarchical clustering.

We evaluate the ability of each model to derive patient representations without relying on true labels during training. For the statistical method, we used a fixed-size concatenated vector of statistical moments across distinct time windows. This methodology was applied without any parameter adjustments, thus requiring no training.

We trained autoregressive deep learning baselines for each dataset using RNN architectures with LSTM or GRU layers, each followed by a fully connected layer. We used PyTorch to implement these architectures and trained them with a learning rate of $10^{-4}$ using the AdamW optimizer to minimize the Mean Squared Error (MSE) loss \cite{adamw}.

We retained the hidden state from the last timestep as the patient representation embedding for both architectures. We observed a gradual decrease in validation loss across the three datasets. This gradual decrease indicates that, at least, some information is inferred for predicting the next timestep values in the autoregressive self-supervised setting.
\section{Results}

\subsection{Patient cohort cluster evaluation}
We generated vector representations of patient trajectories using the statistical and deep learning baselines. 
We trained and evaluated distinct models for each of the three datasets. % , excluding HiRID due to the absence of diagnosis information.  
At inference time, we retrieved statistical and neural embeddings from each model. In our experiments, self-unsupervised training neural baselines outperform, in most cases, the statistical ones, demonstrating that machine learning is capable of learning the dataset's non-linearities more robustly.

\begin{figure}[!ht]
    \centering
    \begin{minipage}[c]{0.02\textwidth}
        % \centering
        \rotatebox{90}{\textbf{v-measure}}
    \end{minipage}%
    \begin{minipage}[c]{0.9\textwidth}
        % \centering
        \begin{subfigure}[b]{\textwidth}
            % \centering
            \includegraphics[scale=0.125]{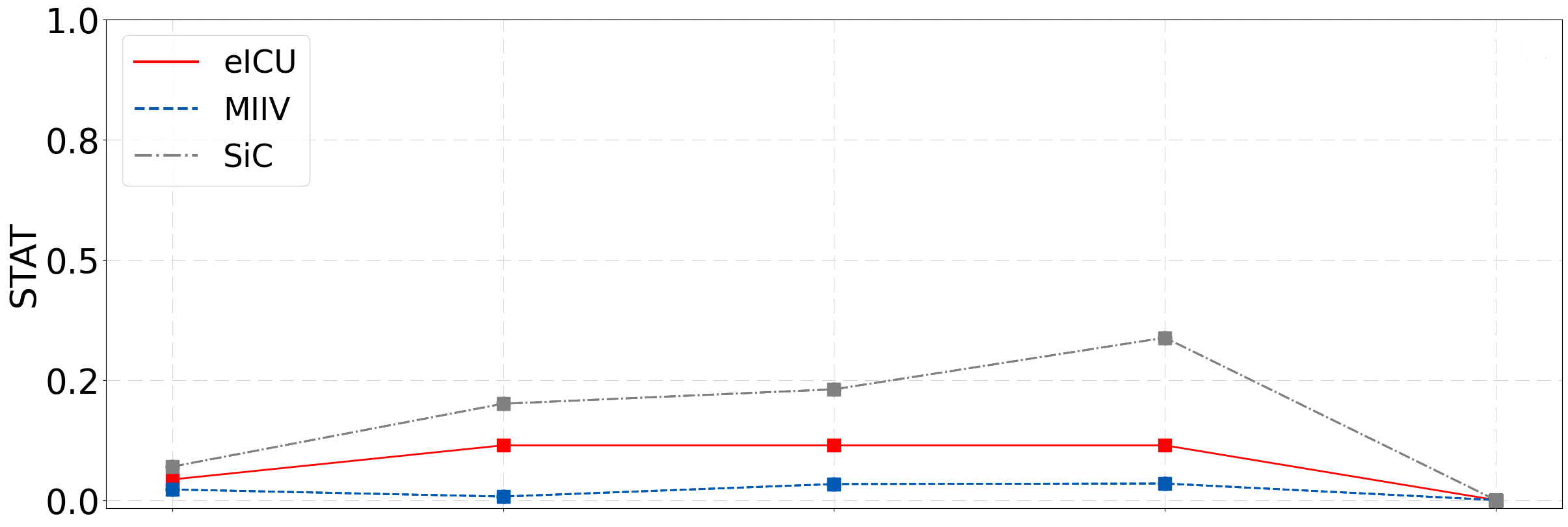}
            \vspace{-0.25ex} % reduce vertical spacing slightly
        \end{subfigure}
        
        \begin{subfigure}[b]{\textwidth}
            % \centering
            \includegraphics[scale=0.125]{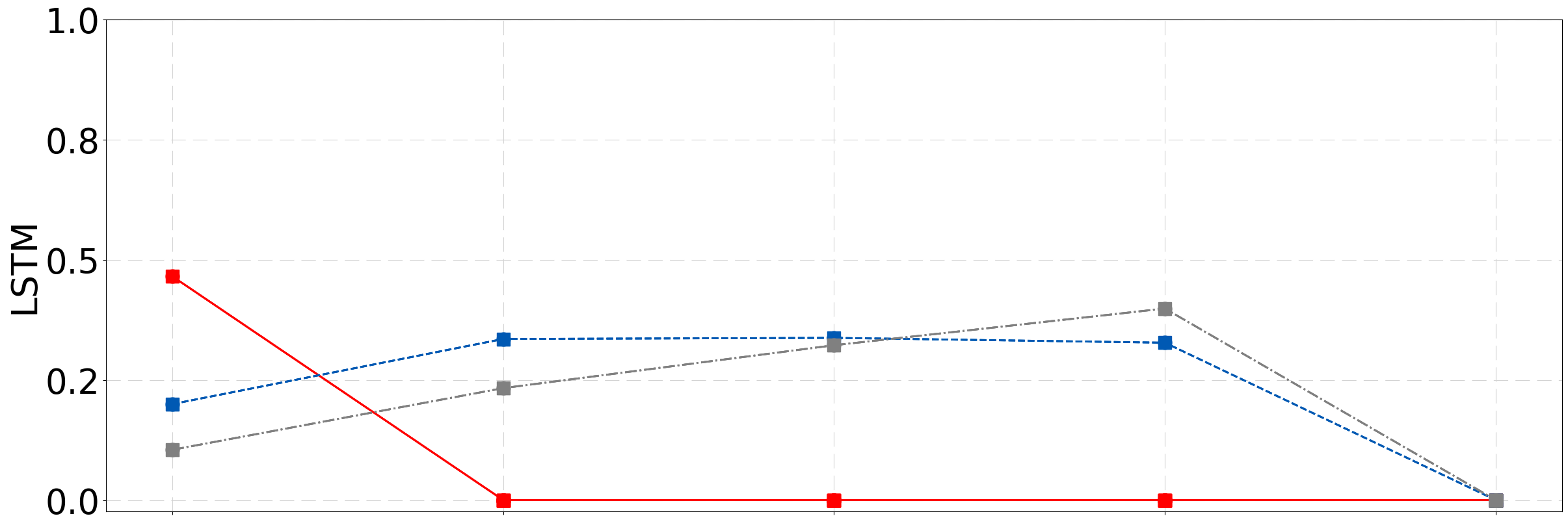}
            \vspace{-0.25ex} % reduce vertical spacing slightly
        \end{subfigure}
        
        \begin{subfigure}[b]{\textwidth}
            % \centering
            \includegraphics[scale=0.125]{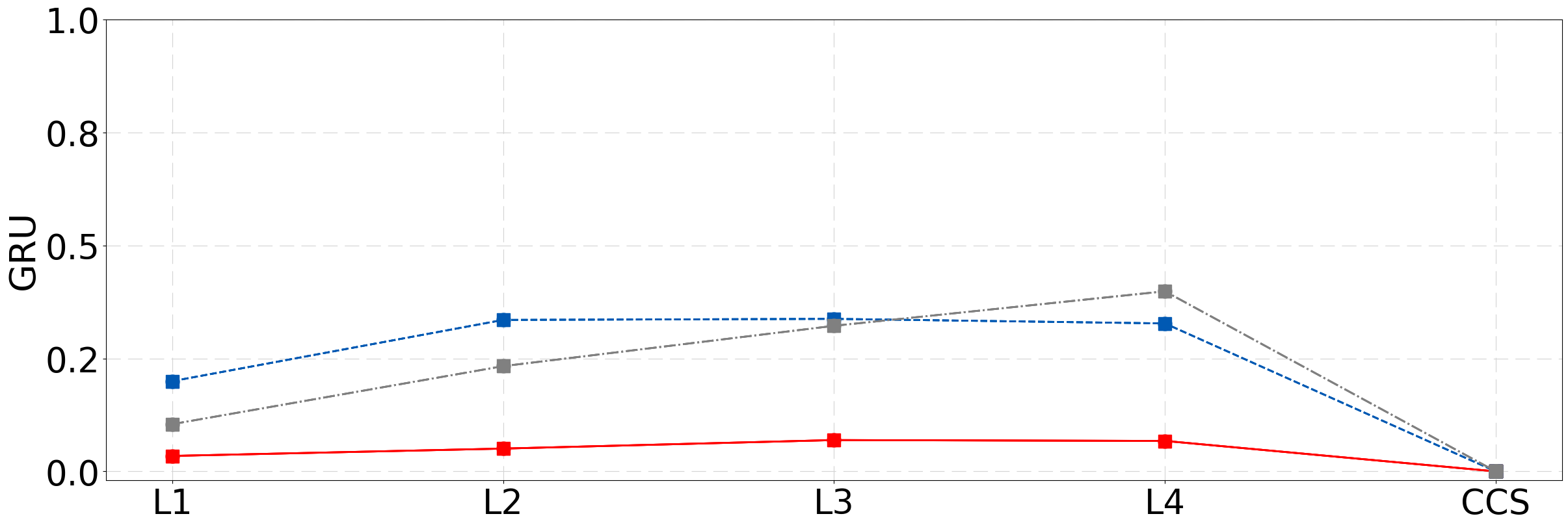}
        \end{subfigure}
    \end{minipage}
    \caption{Clustering evaluation using the v-measure metric comparison for ICD and CCS codes across datasets and problem formulations for STAT, LSTM, and GRU models.}
    \label{fig:vmes}
\end{figure}

Using the neural embeddings from the two machine learning baselines and the STAT patient temporal representations, we compared clustering algorithms based on feature vectors using $k$-Means with varying $k$ values for each clustering task and applying $t$-SNE for dimensionality reduction. Finally, we evaluated the different clustering evaluation formulations (i.e., $L_1-L_4$) with performance indicating the model could rediscover higher hierarchy levels. 
We summarize the results for the v-measure per dataset in Figure \ref{fig:vmes}. 
While both metrics partially rediscovered parts of the ICD hierarchy, the LSTM models significantly surpassed STAT across all problem definitions. We observe an upward trend as the task becomes more complex, aligning proportionally with the number of clusters to be rediscovered, with the notable exception of the eICU dataset for GRU-based architecture.

Finally, regarding performance across datasets, we improved significantly for the eICU dataset, indicating that the LSTM baseline leveraged the high number of features to perform more effectively. On the other hand, model performance on the SiC dataset was worse using LSTM and GRU baselines in the unsupervised; and last, MIMIC-IV had a consistent performance.

\subsection{Hierarchical rediscovery evaluation}

\begin{figure}[!ht]
    \centering
    \begin{minipage}[c]{0.02\textwidth}
        % \centering
        \rotatebox{90}{\textbf{top-1 accuracy}}
    \end{minipage}%
    \begin{minipage}[c]{0.9\textwidth}
        % \centering
        \begin{subfigure}[b]{\textwidth}
            % \centering
            \includegraphics[scale=0.125]{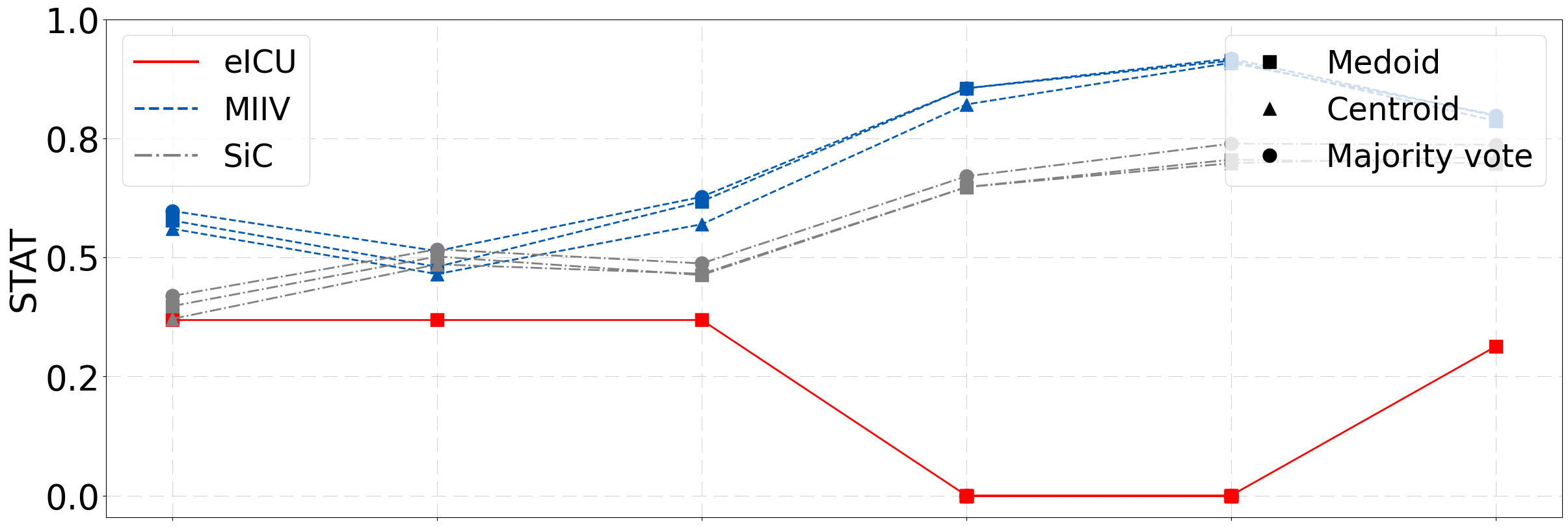}
            \vspace{-0.25ex} % reduce vertical spacing slightly
        \end{subfigure}
        
        \begin{subfigure}[b]{\textwidth}
            % \centering
            \includegraphics[scale=0.125]{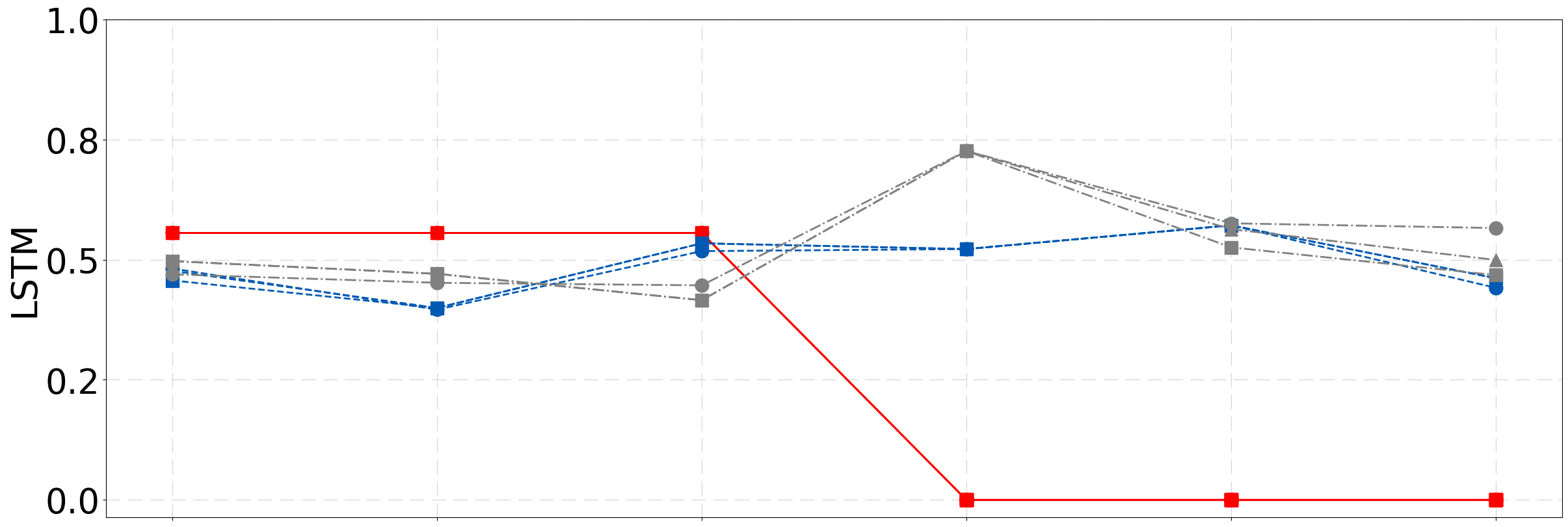}
            \vspace{-0.25ex} % reduce vertical spacing slightly
        \end{subfigure}
        
        \begin{subfigure}[b]{\textwidth}
            % \centering
            \includegraphics[scale=0.125]{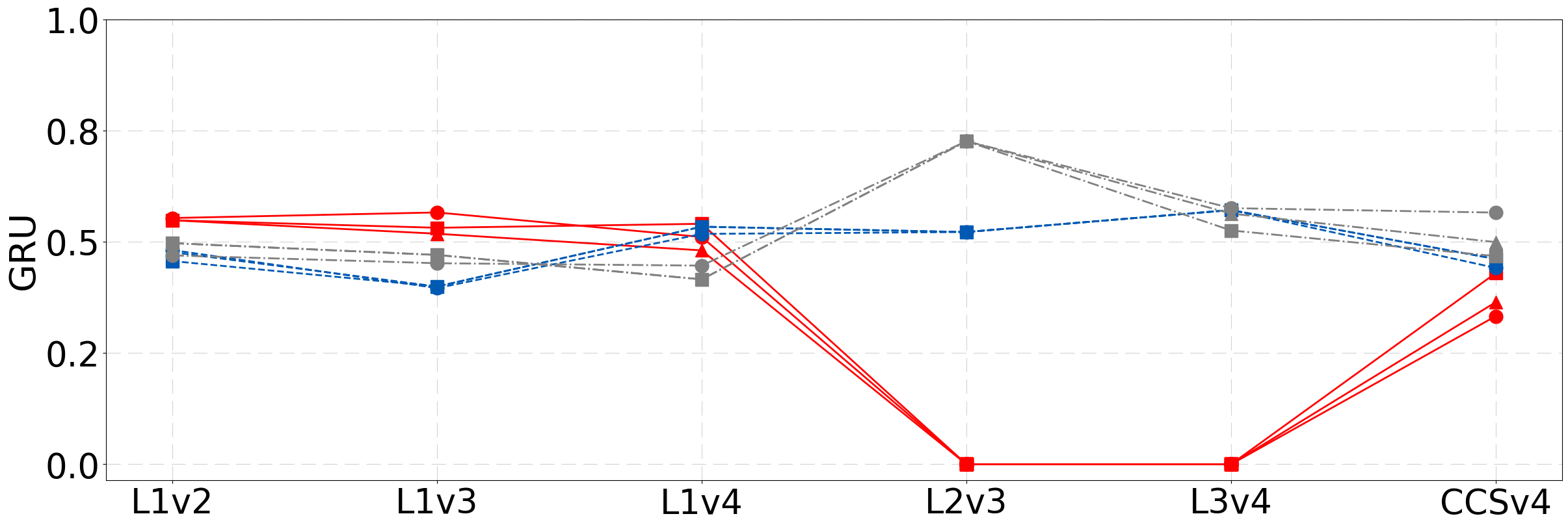}
        \end{subfigure}
    \end{minipage}
    \caption{Evaluation using top-1 accuracy metric comparison across datasets in the hierarchy rediscovery problem formulations using STAT, LSTM, and GRU models.}
    \label{fig:hier_top}
\end{figure}

We averaged the accuracy scores for the clustering problems $L_{k} \xrightarrow{} L_{k+1}$ for the embeddings derived by self-supervised methods, as illustrated in Figure \ref{fig:hier_top}.
We observe inconsistent trends and missing points for the lower levels of the ICD hierarchy, which can be accounted for the smaller number of samples among categories with respect to our criteria for sufficiently large clusters comprising more than 10 samples.

We can retain sufficiently large groups once we relax this assumption to the minimum number to form a distinct cluster in a higher setting. However, they do not qualify for comparison. For all the datasets, we observe an initial trend to discover more effectively the higher levels of our hierarchies. At the same time, performance drops in $L_2 \xrightarrow{} L_4$ and $L_3 \xrightarrow{} L_4$, in the more granular ICD hierarchy level.

\subsection{Cluster label assignment evaluation}
In this setting, we used our training set to assign labels to each cluster. We used the true label of the centroid and medoid cluster points in addition to the majority based to assign a cluster label to each dataset and assess embeddings with respect to the most representative sample of the cohort. 

First, we observe that the performance tasks across tasks is reversed.
We improved performance across datasets significantly for the eICU dataset; the RNN  baselines leveraged the high number of features to perform more effectively. In contrast, the opposite is true for SiC, as illustrated in Figure \ref{fig:hier_top}. 
The majority vote outperformed both metrics across datasets, while the medoid  performed consistently worse.

\begin{figure}[!ht]
    \centering
    \begin{minipage}[c]{0.02\textwidth}
        % \centering
        \rotatebox{90}{\textbf{top-1 accuracy}}
    \end{minipage}%
    \begin{minipage}[c]{0.9\textwidth}
        % \centering
        \begin{subfigure}[b]{\textwidth}
            % \centering
            \includegraphics[scale=0.125]{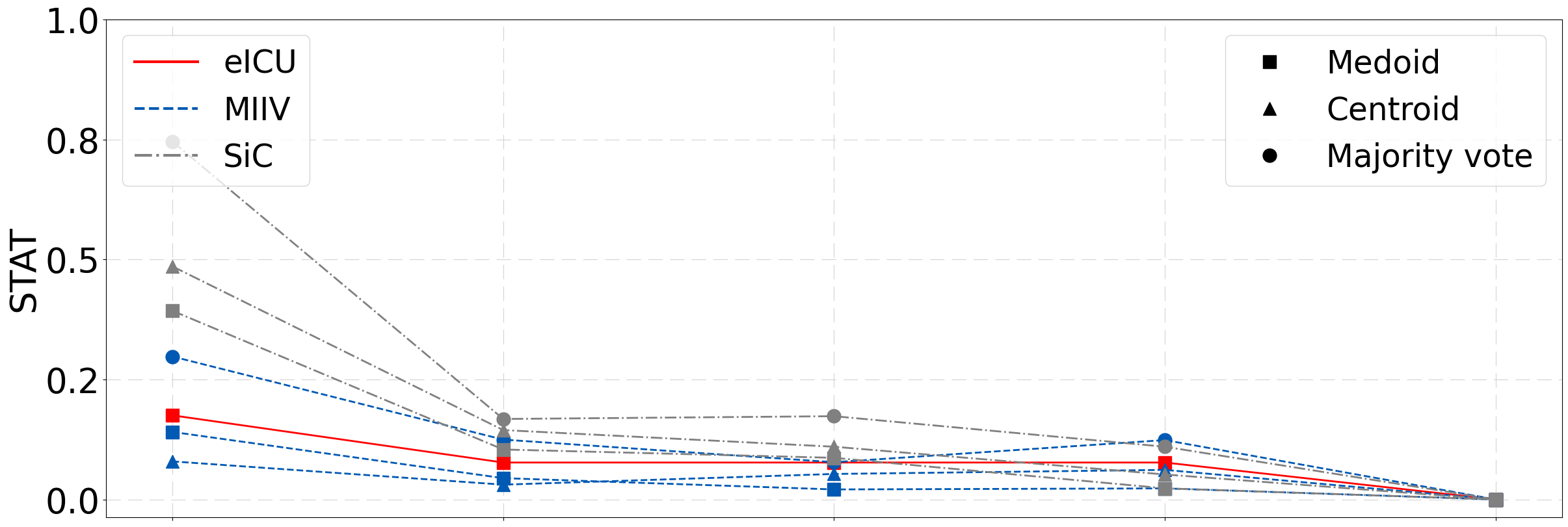}
            \vspace{-0.25ex} % reduce vertical spacing slightly
        \end{subfigure}
        
        \begin{subfigure}[b]{\textwidth}
            % \centering
            \includegraphics[scale=0.125]{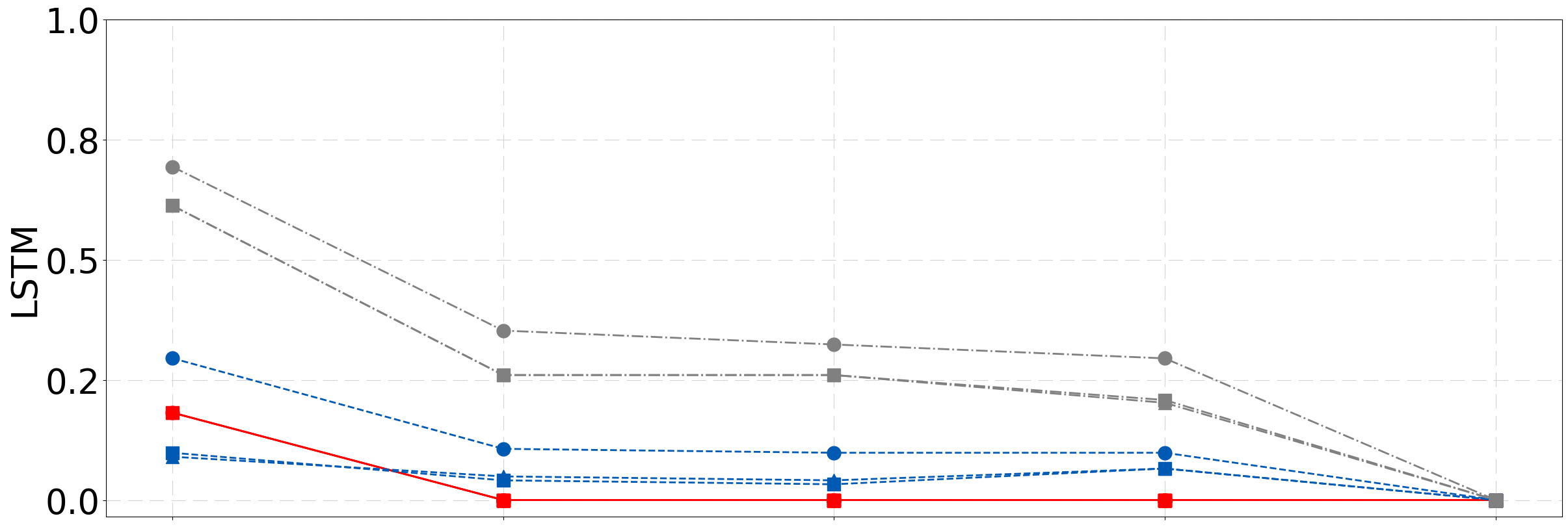}
            \vspace{-0.25ex} % reduce vertical spacing slightly
        \end{subfigure}
        
        \begin{subfigure}[b]{\textwidth}
            % \centering
            \includegraphics[scale=0.125]{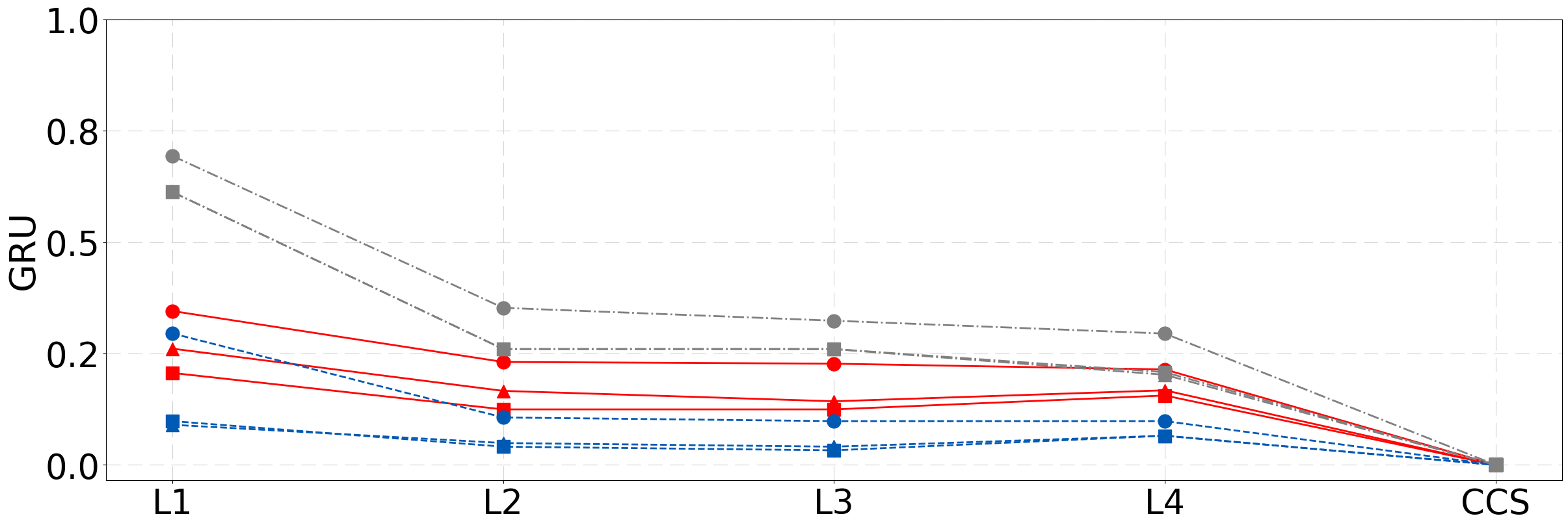}
        \end{subfigure}
    \end{minipage}
    \caption{Evaluation on label assignment strategy on ICD and CCS codes, using top-1 accuracy metric for comparison across datasets using STAT, LSTM, and GRU models.}
    \label{fig:combined_v_measure1}
\end{figure}

Regarding performance across datasets, we observe that some benefited more than others from creating nonlinear vectorial embeddings as illustrated in Figure \ref{fig:combined_v_measure1}. Furthermore, it seems proportional to the number of features. However, for the SiC dataset, the trend appears to be reversed, achieving more accurate performance using the statistical method. Finally, there was unanimous agreement regarding the task's difficulty concerning model performance.

\section{Discussion}
\setlength{\parskip}{0pt} 
Temporal patient representation learning in multicentric ICU datasets presents challenges that influence the performance of unsupervised stratification. Our results reveal insights into representation learning, clustering robustness, and interpretability in unsupervised patient stratification.

One challenge of our approach is that LSTM- and GRU-based models perform optimally on regularly sampled data, whereas ICU records often contain irregular and missing values. While our STAT baseline provides a simpler alternative, its performance lagged behind deep learning models, highlighting the importance of nonlinear temporal patterns for effective patient stratification. Future work could explore transformer-based architectures and adaptive time-series to capture long-range dependencies in EHR sequences.

Another challenge is the skewed label distribution within ICU datasets. ICD codes follow a long-tailed distribution, where common diagnoses dominate while rare conditions are underrepresented. Although our hierarchical clustering approach partially rediscovered ICD subcategories, performance decreased at finer-grained levels ($L_3 \rightarrow L_4$), likely due to insufficient sample sizes. Addressing this imbalance through data augmentation, contrastive learning, or semi-supervised approaches could improve stratification at more granular levels.

In unsupervised settings, extrinsic evaluation remains a challenge. While traditional clustering metrics like the Silhouette Score provided internal validation, our results show that adjusted mutual information (AMI) was the most reliable metric for evaluating alignment with ICD taxonomies. Intrinsic clustering quality does not always correlate with clinical relevance, emphasizing the need for domain-specific benchmarks. Future research could investigate graph-based clustering metrics or clinical validation studies to enhance interpretability.

To improve clinical applicability, we tested three cluster label assignment strategies: centroid-based, medoid-based, and majority vote. Our results show that majority vote produced the most robust cluster assignments, reinforcing that patient cohorts naturally align with common diagnostic categories. However, centroid and medoid-based methods showed potential for capturing subtypes within broader ICD codes, suggesting that hybrid approaches may be valuable in refining clinical ontologies.

\section{Conclusions}
In this work, we introduced ICU-TSB, a reproducible framework for unsupervised patient stratification using temporal representation learning from multicentric ICU datasets. We evaluated statistical and deep learning approaches, comparing LSTM and GRU against a statistical baseline to assess their ability to generate meaningful patient representations. Using ICD and CCS taxonomies, we formulated ICU-TSB, a hierarchical stratification benchmark, allowing models to be evaluated to rediscover structured disease subgroups.

Our experiments demonstrate that patient representations obtained via self-supervised learning significantly outperform statistical baselines in clustering tasks, particularly at higher levels of the ICD hierarchy. Furthermore, we explored the interpretation of the resulting clusters, showing that the assignment of the majority vote label provides the most robust alignment with clinical classifications.

Future work can enhance patient stratification using temporal models (e.g., transformers, diffusion models, graph neural networks) to capture complex EHR sequences, semi-supervised learning to leverage both labeled and unlabeled data, and multimodal integration by fusing clinical notes, imaging, and genomics for richer patient representations. Additionally, graph-based clustering metrics and contrastive learning could improve interpretability.

\paragraph{Acknowledgments}
This work was supported by the Swiss National Science Foundation (SNSF) under grant number 229203 (“AIIDKIT: Artificial Intelligence for Improved Infectious Diseases Outcomes in Kidney Transplant Recipients”); the Sao Paulo Research Foundation (FAPESP -- grants 2024/04761-0, and 19/07665-4); the National Research Council (CNPq 307946/2021-5), and the Coordination for Higher Education Personnel Improvement (CAPES -- grant 001).

\bibliography{aaai25}
\end{document}